%% file: main.tex
\documentclass[11pt,a4paper]{article}
\usepackage[hyperref]{acl2021-templates/acl2021}
\usepackage{times}
\usepackage{latexsym}
\usepackage{graphicx}
\usepackage{hyperref}
\usepackage{multirow}
\usepackage{amsmath}
\usepackage{amssymb}
\usepackage{array,multirow}
\usepackage{booktabs}
\usepackage{color, colortbl}

\usepackage{microtype}

\aclfinalcopy % Uncomment this line for the final submission
\def\aclpaperid{2969} %  Enter the acl Paper ID here

\definecolor{Gray}{gray}{0.9}

\title{Increasing Faithfulness in Knowledge-Grounded Dialogue with Controllable Features }

\author{Hannah Rashkin ~~~~ David Reitter ~~~~ Gaurav Singh Tomar ~~~~ Dipanjan Das\\
Google Research, New York, NY \\
\texttt{\small \{hrashkin, reitter, gtomar, dipanjand\}@google.com} \\
}

\date{}

\begin{document}
\maketitle
\begin{abstract}
Knowledge-grounded dialogue systems are intended to convey information that is based on evidence provided in a given source text. 
We discuss the challenges of training a generative neural dialogue model for such systems that is controlled to stay faithful to the evidence. Existing datasets contain a mix of conversational responses that are faithful to selected evidence as well as more subjective or chit-chat style responses. 
We propose different evaluation measures to disentangle these different styles of responses by quantifying the informativeness and objectivity. 
At training time, additional inputs based on these evaluation measures are given to the dialogue model.  At generation time, these additional inputs act as stylistic controls that encourage the model to generate responses that are faithful to the provided evidence. We also investigate the usage of additional controls at decoding time using resampling techniques. In addition to automatic metrics, we perform a human evaluation study where raters judge the output of these controlled generation models to be generally more objective and faithful to the evidence compared to baseline dialogue systems.
\end{abstract}

\section{Introduction}

Dialogue systems that strive to be informative teachers are difficult to build, despite recent progress in training end-to-end systems that mimic human language at a linguistic level. These systems benefit from vast training data and great representational capacity; yet there are no controls (or training objectives) available that ensure they are truthful. A more limited goal for a system is to be \emph{faithful} to one or more source documents that we implicitly trust. Such a system might help educate users about a particular topic through conversational interaction, or it might augment a task-oriented dialogue system by providing additional information about the process involved in, say, adding a new home automation device. We assume that multi-turn conversational interaction can help a human user learn to retain the new material.

\begin{figure}
    \centering
    \includegraphics[trim={.25cm 0cm .1cm 0cm},clip,width=\columnwidth]{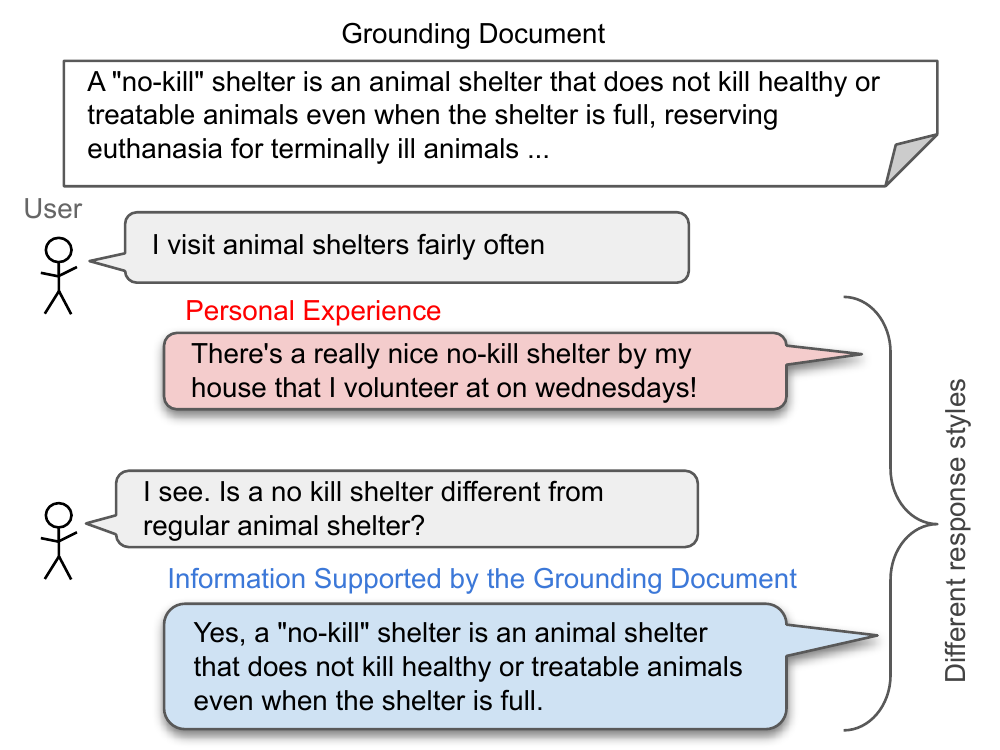}
    \caption{Excerpt from Wizard of Wikipedia \citep{dinan2019wizard} conversation. This grounded dialogue includes responses containing subjective or personal experiences as well as responses sharing information supported by external documents. 
    }
    \label{fig:intro}
\end{figure}

Here, we investigate ways to stay faithful to information from a text document in a conversation. We approach this problem via the task of \emph{knowledge-grounded dialogue}, where a system produces a dialogue response using a piece of \textit{evidence} from a grounding document and a previous conversation history as input (as in Figure~\ref{fig:intro}). Whereas \textsc{PersonaChat}-style tasks \citep{zhang-etal-2018-personalizing} may focus on dialogue systems that are meant to be engaging, this task focuses instead on systems that are meant to be informative, meaning that they only share verifiable information and exclude subjective or invented personal information.  

There are existing knowledge-grounded dialogue datasets (e.g. \cite{ghazvininejad2018knowledge,dinan2019wizard,qin-etal-2019-conversing}) that could be appropriate training resources for such an informative dialogue agent.
However, we observe that these datasets often contain utterances with varying conversation styles and intents, including some utterances that are more informative and some that are chit-chat utterances or subjective commentary.  For instance, in Figure~\ref{fig:intro}, we show an example conversation excerpt from the Wizard of Wikipedia \citep{dinan2019wizard} training set.  While some utterances are supported by the grounding documents (the second response), others include personal experiences and observations (as in the first response).  Because of this mix of conversations styles, we cannot ensure that models naively trained on this data will learn to generate only faithful, informative utterances. 

In order to avoid this issue, one could collect new datasets where the responses are more explicitly constrained by the evidence, but this could be quite expensive and may be challenging to implement. Instead, in this paper, we propose an alternate approach: we adapt techniques from controllable text generation in order to train dialogue models that learn to disentangle these conversation styles within the data and can be controlled at generation time to produce more grounded responses.

We propose using multiple evaluation measures that are relevant to the faithfulness of a response and use these to control the output of two commonly used seq2seq models (GPT-2 \cite{Radford2019LanguageMA} and T5 \cite{t5ppr}). We investigate two methods for adding controllability. First, we integrate control code features based on the evaluation measures as special tokens prepended to the seq2seq input, drawing inspiration from domain-based control codes methods \citep{CTRL}. These special tokens are created using information about the gold response at training time, but are set to maximize the groundedness of the responses at generation time.  Second, we implement a form of resampling that directly restricts the output to satisfy the proposed evaluation measures.

In order to inspect the faithfulness and style of the responses, we use automatic evaluations (including BLEU and the evaluation measures described) and human evaluations that are designed to focus on the degree to which the response is faithfully representing information from the evidence. Our results show that using these controllable generation techniques can improve the perceived faithfulness and objectivity. We also show that the proposed evaluation measures correlate with the human judgements, indicating that these are appropriate measures for gauging specific aspects of groundedness. Lastly, we conclude the paper with some discussion of examples and possible trade-offs.

\section{Task}
We introduce a sub-task of knowledge-grounded dialogue where a dialogue agent is intended to be informative and must not share {\it hallucinations}, which we define here as any information that is neither inferrable from nor directly stated by external documents.  In this task, a system is given evidence from a document (or documents) and a conversation history and must produce a response that is both faithful to the evidence and also natural within the context of the previous conversation utterances. 
Because this task focuses on being informative to a user, the agent is not allowed to share unsupported or subjective information (this includes invented personal traits - e.g. ``I love dogs, too!'').  Additionally, it is not sufficient to be purely extractive as information from the evidence may need to be re-phrased to be a conversationally appropriate response (e.g. if a user asked a question that is inferrable from the evidence but not directly stated).

To simplify the task for this paper, we assume that an appropriate evidence span, $e$, has already been labelled.  We therefore study how to generate an appropriate response $y$ given the previous conversation history $x$ and a chosen evidence $e$ as input.

\subsection{Evaluation measures}
\label{sec:measures}
Our goal is to design a dialogue model that is more faithful and objective in how it relays evidence.  We propose using a series of evaluation measures to estimate whether a response is (1) written in an objective voice, (2) not sharing extra information that is not in the document and (3) entailed by the grounding evidence.
In the modeling section (Sec. ~\ref{sec:methods}), we describe how we incorporate these measures into a controllable generation framework.

\paragraph{Objective Voice}
One form of hallucination is when a dialogue agent might share personal stories or opinions.  It is common for dialogue agents to learn this behavior as many dialogue datasets contain instances of personal chit-chat even if the task is aimed at grounded language. We estimate objective voice as a binary variable based on the presence of first person singular pronouns detected using a word list.
 
\paragraph{Lexical Precision}
We also want to ensure that the response is not adding extra information from what's in the selected evidence.  To estimate this, we measure the precision of the unigrams in the response with respect to the evidence.  A high value indicates that most of the words in the response are contained somewhere in the evidence. We use this measure because it is relevant to grounding precision scores in previous work \cite{Tian2020ResponseAnticipatedMF} and because it can reasonably gauge how extractive the response is, but one drawback of this measure is that it is based on lexical features which may not reflect semantic differences in the information being shared (e.g. dropping the word `not' may yield high lexical precision but a very different semantic meaning from the original evidence). We leave investigation of more semantic-oriented measures of the precision of information to future work.

\paragraph{Entailment}
Lastly, we want to encourage the model to produce a response that is semantically entailed by the source document.  We use a state-of-the-art natural language interference (NLI) model (Roberta trained on MNLI \cite{roberta:mnli}) to estimate if a response is entailed by the evidence.\footnote{We aggregate neutral and contradiction as ``non-entailing'' because we care mainly about detecting entailment rather than the distinctions between the other two standard NLI categories.}

\section{Data}
\emph{Wizard of Wikipedia} \cite{dinan2019wizard} is a recent, large-scale dataset of multi-turn knowledge-grounded dialogues between a ``apprentice'' and a ``wizard'', who has access to information from Wikipedia documents.  The wizard labelled evidence spans within documents for each utterance they made.  Additionally, the development and test sets are split into two portions depending on if the conversation is about a topic that was seen or unseen in the training data.  We use the gold-labelled evidence as input to the model in order to focus on improving the quality of generating responses given such evidence and the previous dialogue history. We also focus on only modeling the utterances by the ``wizard'' in the cases where they are responding to the ``apprentice''.  We include data statistics in  Table~\ref{tab:datastats} and an example conversation excerpt in Figure~\ref{fig:intro}.

\begin{table}[]
    \centering
    \small
    \begin{tabular}{cc}
    \toprule
        \multicolumn{2}{c}{Wizard of Wikipedia \citep{dinan2019wizard}}  \\
        \midrule
         &\# Wizard responses \\
         Train data & 73571\\
         Dev (seen topics) &  3905\\
         Dev (unseen topics) & 3898\\
         Test (seen topics) & 3842\\
         Test (unseen topics) & 3902 \\
         \midrule
         \multicolumn{2}{c}{Training responses}\\
         \% with first person & 44\%\\
         Avg. lexical prec. wrt evid. & 0.43 \\
         \% predicted entailed & 23\% \\
         \bottomrule
    \end{tabular}
    \caption{Data statistics from the Wizard of Wikipedia dataset.}
    \label{tab:datastats}
\end{table}
\begin{figure*}
    \centering
    \includegraphics[trim={.6cm 1.5cm .6cm 0cm}, clip, width=.9\textwidth]{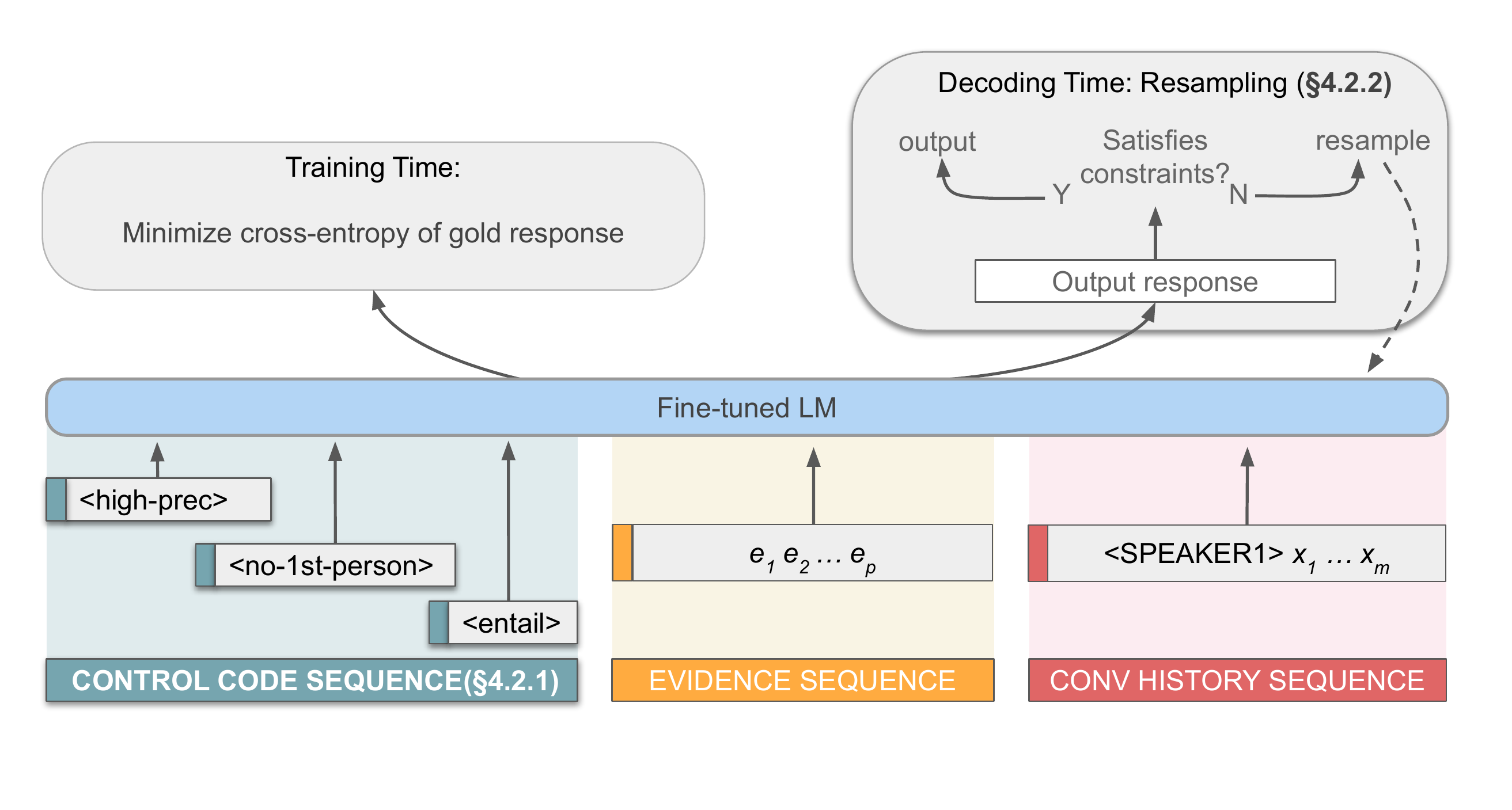}
    \caption{Modeling Figure: In our modeling framework, a large pre-trained language model is used to encode the evidence and conversation history and produce a response.  We incorporate additional tokens (i.e. control codes) to train the model to recognize differences between types of utterance that are more or less grounded to the evidence.  At decoding time, we also investigate the significance of using resampling methods.}
    \label{fig:model}
\end{figure*}

We note that even though Wizard of Wikipedia is a knowledge-grounded dataset, there are many utterances that also include information external to the evidence (as noted in Figure~\ref{fig:intro}). Many conversation turns relay evidence while also embellishing with chit-chat, opinion sharing, or interlocutors' own intuitions and world knowledge. This is because this dataset was collected by asking human crowdworkers to converse with each other, and it is natural for humans to embellish and personalize their conversations even when discussing a document. 
Yet, for our goal of training informative dialogue agents, we need to train models that only relay information that is found in the evidence.  

In order to avoid collecting new data, which is costly and challenging, we investigate how to train models with this data while discouraging them from hallucinating extra information that cannot be confirmed in the evidence. One way to deal with this challenge might be to only train with the portions of the data where the response is highly grounded by the evidence.  However, in our calculations (bottom of Table~\ref{tab:datastats}), we find that as much as 44\% of training set responses are in first person and only 23\% of responses are predicted to be entailed by the evidence, which indicates that a large portion of training data would have to be excluded. Instead, our paper proposes a modeling technique in which we incorporate different input features denoting different conversational styles.  We can then train the model in a way that learns to use these features to disentangle the differences between utterances that are more faithful to the evidence vs. other types of utterances.

\section{Modeling}
\label{sec:methods}
We investigate how to add controllable features to a large neural dialogue model in order to constrain the amount of hallucinated text while also taking advantage of the underlying fluency of a large end-to-end neural model.

\subsection{Generation Model} As our underlying dialogue model, we use neural seq2seq architectures -- T5 \cite{t5ppr} and GPT-2 \cite{Radford2019LanguageMA}, which are architectures used in state-of-the-art dialogue systems (e.g. DialoGPT \cite{dialogpt}). We fine-tune these models on our grounded dialogue dataset.  The input to the model is a sequence of evidence tokens $e_1...e_p$ and a dialogue history which we treat as a sequence of tokens $x_1...x_m$ where the utterances are delimited by the speaker ID (either \texttt{<speaker1>} or \texttt{<speaker2>}). For the GPT-2 model, we also include special token-type embeddings that are added to the byte-pair embedding tokens and position embeddings.  The token-type embeddings denote the segments of the input that belong to the evidence and the two different speakers.
We train the model to produce the next conversation utterance $y_1...y_n$ by minimizing the cross-entropy:
\begin{equation}
    \mathcal{L}_{CE} = -\frac{1}{n} \sum_{i=1}^n \log p(y_i | y_{<i}, x, e)
\end{equation}

\paragraph{Caveats of generative language models}
As noted by the documentation with the GPT-2 release, we lack a complete understanding of language models' robustness and worst case behaviors.
Even though training data for GPT-2 and T5 have been carefully selected, these large datasets may contain sources with unfair distributions and factual inaccuracies, and thus the models and the resulting generated synthetic data may have inherited these biases. Additionally, the output generated by these models may only succeed in being superficially similar to human-written text or dialogue turns. 

\subsection{Adding controllable generation}
We describe two methods of adding controllability to the dialogue models to enhance the groundedness according to the evaluation measures from Sec.~\ref{sec:measures}.  First, we incorporate control features into the input of the model.  Second, we describe additional decoding-time techniques using resampling.

\subsubsection{Control Code Features}
\label{methods:soft}
We add control features as a way of encouraging the underlying language model to disentangle different conversations styles at training time.  We implement this using the control code approach previously introduced in CTRL \cite{CTRL}.  First, we use the measures introduced in Section~\ref{sec:measures} to create control feature tokens based on how much of the content of the response is grounded in the gold labelled evidence. The control feature tokens $c_1...c_n$ are prepended to the other tokens. The training objective therefore becomes:
\begin{equation}
    \mathcal{L}_{CE} = -\frac{1}{n} \sum_{i=1}^n \log p(y_i | y_{<i}, x, e, c)
\end{equation}

At training time, we set control feature tokens based on measures of entailment, lexical precision, and objective voice of the gold response. At decoding time, control codes are set to the desired valued for these qualities (high entailment, high lexical precision, objective voice).

\paragraph{Objective Voice}  
In order to encourage the model to be only relaying objective information from the evidence, we include a control code for whether or not the utterance contains first-person pronouns (\texttt{<first-person>},\texttt{<no-first-person>}). At decoding time, we always use the \texttt{<no-first-person>} control token.
 
\paragraph{Lexical Precision}
We measure the lexical precision of the response with respect to the evidence, splitting the training utterances into three terciles (high, medium, and low). We map the terciles to control codes to denote the precision level (\texttt{<high-prec>},\texttt{<med-prec>}, and \texttt{<low-prec>}). At decoding time, we always use \texttt{<high-prec>}.

\paragraph{Entailment}
We add control codes for the output of the NLI classifier (\texttt{<entailed>},\texttt{<non-entailed>}). At decoding time, we always use \texttt{<entailed>}.

\subsubsection{Controlled resampling}
Whereas the control code method implicitly teaches the model to use different styles, some applications may require more direct control over the model output.  Additionally, there may be situations where a dialogue system cannot be re-trained.  We therefore also investigate a method of implementing more direct control at  decoding time. We experiment with a resampling method that continues to sample responses until one is found that satisfies the evaluation measures (high lexical precision, objective voice, and predicted entailment). To save on computational efficiency, we use a cut-off to avoid resampling more than $d$ times.

\input{ablation_tab}
\section{Experiments}
We perform experiments using automatic metrics and human judgments to evaluate the effectiveness of the proposed controllable dialogue system and its various components.

\subsection{Set-up}
We use the HuggingFace library \cite{huggingface} versions of GPT-2 and T5. We select training hyperparameters based on cross-entropy of the development set.  We use a learning rate of $8E-5$ and maximum gradient norm of {$1$, $3.5$} for {GPT-2, T5} respectively with ADAM to minimize the training loss (with 200 warm-up steps). If the total sequence length is greater than 1024, we truncate the previous conversation turns until the sequence is short enough. We train for three epochs for all models.  For decoding, we use nucleus sampling \cite{holtzman2020} with $p=0.6$ and a minimum generation length of five tokens (based on better BLEU performance with the development set). In our experiments with resampling, we arbitrarily set $d=10$.

\input{automet_tab}

\subsection{Metrics}
We use both automatic metrics (Sec.~\ref{sec:auto:abl} and \ref{sec:auto:full}) and human ratings (Sec.~\ref{sec:humanstudy}) to better understand performance of our model and the effect of controllable features. 

First, we use BLEU to compare the model output to a gold reference.  While BLEU gives a general sense of the fluency, there are drawbacks to word-overlap metrics for evaluating open-ended generations like dialogue \citep{hownottoeval}. 
Additionally, comparing to a gold reference answer fails to measure the underlying question we hope to interpret: whether the response is more objective and grounded to the evidence.  Therefore, we also evaluate the output using the proposed evaluation measures from Section~\ref{sec:measures}. In addition to  lexical precision, we also report the lexical recall of words from the evidence.  

But, the controllable models are controlled using the same evaluation measures, so we expect that these models may have an advantage in these metrics.  Thus, we rely more on human evaluations (Section~\ref{sec:humanstudy}). We ask humans to evaluate the quality along multiple aspects including whether the response is fluent, relevant, supported/faithful, and objective.

\subsection{Ablation of Control Code Features}
\label{sec:auto:abl}
First we conduct an ablation study to investigate the effects of each individual control code feature being used as model input.  Table~\ref{tab:ablation} shows the results on the seen topics portion of the Wizard of Wikipedia development set. Unsurprisingly, each control feature generally helps in improving on the measure that was used in its training. However, we also find, more generally, that each type of control code feature does improve over the base model on all metrics.  
Results also show that using all control code features together generally improves the performance across the automatic metrics.

\begin{table*}[]
    \centering
    \begin{tabular}{r|llll}
    \toprule
        Model & Fluency & Relevance & Faithfulness & Objectivity \\
    \midrule
        E2E model \cite{dinan2019wizard} & $5.00$	& $4.61^{\ }$&	$2.48^{\text{***}}$&$2.26^{\text{***}}$\\
        dodecaDialogue \cite{dodeca} & $4.99$ & $4.73^{\ }$ & $4.37^{\text{*}}$ & $3.92^{\text{**}}$\\
        T5 (none) & $5.00$ & ${\bf 4.84^{\ }}$ & $3.66^{\text{***}}$ & $2.82^{\text{***}}$\\
        T5 (control codes)& $5.00$ & $4.66^{\text{*}}$ & ${\bf 4.64^{\ }}$ & ${\bf 4.53^{\ }}$\\
        T5 (resampling)& $4.99$ & $4.77^{\ }$ & $4.14^{\text{***}}$ & $3.82^{\text{***}}$\\
        T5 (both)& $5.00$ & $4.82^{\ }$ & ${4.42^{\text{*}}}$ & ${4.30^{\ }}$\\
    \bottomrule
    \end{tabular}
    \caption{Human Quality Ratings: $^{\text{*}}$,$^{\text{**}}$ ,$^{\text{***}}$ indicates that this result is significantly different from the best result in that column (bolded) with p-value $<0.05$, $<0.01$, $<0.001$ respectively.}
    \label{tab:humanstudy}
\end{table*}

\subsection{Automatic Metric Results on Test Set}
\label{sec:auto:full}
We show results on both portions of the Wizard of Wikipedia test set in Table~\ref{tab:fullresults}.  As baselines, we use finetuned GPT-2 and T5 without any controllable features or resampling.  We also include results the end-to-end generative model (E2E) with gold knowledge that was introduced in the original Wizard of Wikipedia paper \cite{dinan2019wizard} and the model in the follow-up work on dodecaDialogue \cite{dodeca}.  These are transformer-based architectures that use the evidence and conversation history as inputs but do not explicitly control the model to be more faithful to the input.  In general, we find that models with pre-trained or multi-task training set-ups (dodecaDialogue, GPT-2, and T5) have relatively consistent performance across both the seen and unseen topic partitions of the test set, indicating that these models can generalize fairly well to unseen topics. 

Results generally show improvements over the baselines when using control codes.  By additionally using resampling at decoding time, we see further improvements, though resampling is not as effective on its own.  One explanation why resampling is not as effective is that it may be unable to find a satisfactory response within $d$ resampling turns, particularly if the underlying model has not been already trained in a controllable set-up.  Supporting this, we find that different choices of $d$ has more of an impact on performance with the ``just resampling'' model than with the ``control code + resampling'' model.

The controllable T5 models generally outperform all of the other models in terms of the metrics from Section~\ref{sec:measures}. This may not be so surprising since these models are using the same metrics for control inputs at training time.  The dodecaDialogue model outperforms our best model variant in the BLEU and recall metrics, but this may also be related to the longer average token length of output of that model (19 tokens on average) in comparison to our model (16 tokens on average).  In order to get a more conclusive understanding of the performance differences, we perform a human evaluation study, described below.

\subsection{Human Evaluation}
\label{sec:humanstudy}
We use human evaluations to gauge performance across multiple aspects of quality.  One aspect which we focus on is how much the information in the responses is grounded in the evidence, which we consider to be a strong requirement for this task. But, there are also other complementary aspects of response quality that are important (e.g. being appropriate to the conversational context).
Therefore, we ask raters to judge a random subsample of model responses from the test set in terms of four qualities: fluency (how understandable and proficient the language is), relevance (whether it is an appropriate reply to the conversation history), faithfulness (whether the reply is fully supported by the evidence), and objectivity (whether the reply is fully objective, rather than sharing personal feelings or experiences).\footnote{The exact phrasing of the questions given to human raters is in the appendix.}  

We subsample examples from the seen topics test set, using 100 examples per model variant with 3 human raters per example.   In order to give raters more flexibility, they are asked to rate each quality on a Likert scale from 1 (low quality) to 5 (high quality).  We measure the agreement for each of the four qualities separately using Krippendorff's alpha and find that the agreement (0.8, 0.91, 0.88, 0.96 respectively) is reliably high.  

In Table~\ref{tab:humanstudy}, we include the averaged results from the human study.  We provide asterisks in every case where a metric is significantly different from the best result (bolded), as found with Welch's t-test.   By adding the control code features and resampling, we do not see a drop in the fluency, which is similarly high across all of the models.  In fact, we see that most of the trade-off is between the relevance of the response vs. the faithfulness and objectivity.

Our results show the faithfulness and objectivity of the T5 models with control codes is significantly higher than in the uncontrolled models (top three rows). This is a promising indication that adding these controllable features significantly steers the generations towards making more grounded, objective responses, with only a slight decrease in relevance.  Including resampling is not as effective in promoting faithfulness and objectivity as the control codes, though more faithful and objective than the base T5 model.   By using both control codes and resampling (bottom row), the T5 model is able to achieve nearly the same level of faithfulness and objectivity as with just using control codes, but with higher relevance subscores.

For the full set of annotated examples, we also find that the human scores for faithfulness and objectivity correlate with measurements from the evaluation measures that we described in Section~\ref{sec:measures}.  For instance, the absence of first person strongly correlates with higher objectivity according to human raters (Pearson r value of 0.8 at p value $<0.001$).  Lexical precision and entailment measures both strongly correlate with human perceptions of faithfulness and objectivity, as well.\footnote{The appendix includes a full table of correlation coefficients}  This confirms that the evaluation measures that we propose using as controls can be  appropriate estimates for how humans might perceive the groundedness of a response.   However, these metrics do not correlate to relevance or fluency. Based on these observations, it seems that these measures can be useful to gauge the general groundedness of the response but should still be viewed in tandem with other quality scores to get a more holistic understanding of performance.

\input{output_examples_excerpt}

\subsection{Qualitative Examples}
In Table~\ref{tab:gen:ex}, we highlight some examples of model output (we also provide additional examples in the appendix). The responses in the controllable models tend to be more concise in relaying information from the evidence.  In the first example, the controllable model only shares information that is entailed by the evidence, excluding extra information about spices that is not easily verifiable within the document.  

This may also come with a slight trade-off with the relevance of the replies, as in the second example where the response - while more faithful to the evidence - is not quite as pertinent to the previous conversation turn. Similarly, in the third example, the full model is faithfully citing the evidence but is too extractive to the extent of including irrelevant details. 
In the last example in Table~\ref{tab:gen:ex}, both the models make the same error where they incorrectly give an affirmative answer to the user's question about George Foreman even though they both identify Michael Boehm as the correct inventor (a better answer would be ``No, it was Michael Boehm.''). This example is challenging because the answer to the user's question is not directly stated in the evidence and requires extra inference rather than just extracting relevant words. 
To address these challenges, one area for future work may be investigating approaches that combine extractive and abstractive generation methods to be more deliberately selective about which portions of evidence are being used and how they are integrated with information about the conversational discourse.

\section{Related Work}
\paragraph{Knowledge-Grounded Dialogue}
There has been significant prior work in tasks for designing dialogue agents that are grounded by document knowledge \cite{dinan2019wizard,qin-etal-2019-conversing,ghazvininejad2018knowledge,Tian2020ResponseAnticipatedMF,topicalchat,moghe-etal-2018-towards}.  Some of these works investigate retrieving appropriate evidence \cite{Lian2019LearningTS,Meng2020DukeNetAD, kim2020sequential}, while we assume that a piece of evidence has already been retrieved and focus instead on how to craft generations that are more faithful to it.  Our work is also novel in investigating controllable generation as one way of disentangling evidence-based utterances from more subjective utterances that may be present in the training data.

\paragraph{Controlling hallucinations in text generation}
There is a body of work that has previously studied methods for integrating evidence in natural language generation tasks, with a focus on reducing hallucinations.  Many of these works focus on other generation tasks such as summarization \cite{absum:factuality,zhao2020reducing,Cao2018FaithfulTT,falke-etal-2019-ranking} or data-to-text generation \cite{Puduppully2019DatatoTextGW}.  We investigate how the problem of reducing hallucinations can be applied to the task of knowledge grounded dialogue.  Similar to our approach, \citet{Filippova2020ControlledHL} also uses control codes to reduce hallucinations but focused instead on data-to-text generation tasks.

\paragraph{Controllable Text Generation}
In order to control the faithfulness of responses, we draw on techniques from controllable text generation tasks.  Most relevant is the development of control-code-style input tokens such as in CTRL \citep{CTRL} or the LFT model of \citet{niu-bansal-2018-polite}.  Others have used decoding-time re-ranking \citep{falke-etal-2019-ranking} to constrain the outputs in a way that is similar to our resampling method. 
Controllable generation has also been used previously with open-ended dialogue data
\cite{see2019convo} to improve qualities such as the engagingness; however, our work focuses on knowledge-grounded dialogues aiming to increase the faithfulness of the replies. Recently, \citet{wu2021controllable} used control phrases as controllable inputs to decrease hallucination as a form of content planning. We similarly use controllable features to reduce hallucinations in knowledge grounded dialogues, but our model uses stylistic measures which can be seen as complementary to content planning.

\section{Conclusion}
In this paper, we investigate how to design knowledge grounded dialogue systems that are less prone to including hallucinations or subjective information. We discuss three evaluation measures related to the groundedness of the response and discuss two methods for integrating these metrics into a controllable dialogue system. We demonstrate that this controllable dialogue system is able to produce responses that are perceived by humans to be more objective and faithful to document-based evidence.

\section*{Acknowledgements}
We would like to thank Slav Petrov and Ankur Parikh as well as the anonymous reviewers for their insightful comments and feedback.  We also thank Nouha Dziri for sharing code and data resources. We additionally thank Ashwin Kakarla and his team for helping with human annotations.

\section*{Impact Statement}
In this paper, we study the problem of encouraging knowledge grounded dialogue agents to be more faithful in generating information from trusted documents.  The controllable models and evaluation measures proposed in this paper could benefit general dialogue applications by constraining their output to only discuss information that is verifiable, which could ensure that these systems are more trustworthy. This could be valuable in a wide range of applications such as educational or information-seeking dialogue settings where the user needs to be given accurate information. As with other conditional generation models, this could also pose a risk if these models were misused by conditioning on evidence from unreliable resources.  In our work, we mitigate this risk by carefully considering the source of our evidence and how it was curated. Before applying these models, others should similarly take into consideration whether their evidence sources are reliable and unbiased.

\bibliographystyle{acl2021-templates/acl_natbib}
\bibliography{main,acl2021-templates/anthology.bib}

\clearpage
\appendix
\section{Appendix}
\label{sec:appendix}
\input{appendix}

\end{document}

%% file: ablation_tab.tex
\begin{table*}[t]
    \centering
    \begin{tabular}{r|cccc|c|cc|c}
    \toprule
        & \multicolumn{4}{c|}{BLEU} & Objectiv. & \multicolumn{2}{c|}{Overlap w.r.t. Evid} & Entail\\
         Control Codes& B1 &B2 &B3 &B4 &\% N1P & Prec & Rec & \% Entail\\
    \midrule
         GPT-2 + No Control & 27.6 &  12.5 &  8.1 &  6.0  & 49.9& 56.4 & 48.4 &  34.9 \\
         + Objective & 26.4 &  12.8 & 8.6 &  6.4 &98.1& 62.3 & 50.5 & 50.1\\
         + High Lex Prec & \textbf{29.9}& \textbf{15.1} & \textbf{10.4}&\textbf{7.9}&63.4&70.6  & \textbf{60.3} & 51.9\\
         + Entailment & 27.3 &   13.9 & 9.4  & 7.1  &80.5& 72.6 & 56.2  &69.7\\
         + All    & 27.9  & 14.8 & 10.2 & 7.7 &\textbf{99.4} & \textbf{76.6} & \textbf{60.3} & \textbf{72.3}\\
    \midrule
    
         T5 + No Control & 28.6 &  14.4 &  9.7 &  7.3 & 49.8& 63.9 & 50.1 &  44.4 \\
         + Objective & 27.4 &  14.8 & 10.3 &  7.9 & 99.4 & 70.7 & 53.4 & 65.1\\
         + High Lex Prec & \textbf{29.6}& \textbf{15.9} & \textbf{11.1}&8.4& 63.3&76.1  & 59.8 & 60.7\\
         + Entailment & 27.8 &   15.3 & 10.6  & 8.1&80.3  & 77.7& 57.3  &80.0\\
         + All    & 27.4  &  15.5 & 11.0 & \textbf{8.5} & \textbf{99.9}& \textbf{84.2} & \textbf{60.7} & \textbf{89.4}\\
         \bottomrule
    \end{tabular}
    \caption{Ablation Study: The effects of using different types of control codes for generation on the Wizard of Wikipedia seen topic development set.  In addition to BLEU, we measure objective voice by the percent of replies where there is no first person (N1P).  We also measure the lexical precision and recall with respect to the words from the evidence. Lastly we compute the percentage of responses that are predicted to be entailed according to an MNLI classifier.}
    \label{tab:ablation}
\end{table*}

%% file: automet_tab.tex
\begin{table*}[t]
    \centering
    \small
    \begin{tabular}{r|ccccc|ccccc}
    \toprule
        & \multicolumn{5}{c|}{Seen Topic} & \multicolumn{5}{c}{Unseen Topic}\\ 
        &BLEU& N1P &\multicolumn{2}{c}{ w/ Evid.} & NLI & BLEU & N1P &\multicolumn{2}{c}{ w/ Evid.} & NLI\\ 
        Model & B4 & \% & Prec & Rec & \% & B4 &  \%&Prec & Rec & \%\\
    \midrule
        E2E model \cite{dinan2019wizard} & 1.5 & 48.0& 47.9 & 30.4 & 29.3 & 0.3 & 37.6& 33.2 & 21.7 & 9.5\\
        dodecaDialogue \cite{dodeca} & \textbf{10.0} & 78.3& 81.1 & \textbf{67.7} & 70.3 & \textbf{ 9.7} & 77.7& 81.3 & \textbf{ 66.5} & 70.6\\\hline
        GPT-2 (none) & 6.2 & 50.9& 56.1 & 49.4 & 34.2&5.7 &52.1& 56.4 & 48.3 & 34.2\\
        GPT-2 (control codes) & 7.8&99.3& 76.6 & 61.5 & 73.8&7.6 &99.2& 77.0 & 60.3 & 74.0\\
        GPT-2 (resampling) & 7.6&75.1& 70.4 & 57.7 & 71.4&7.2 & 76.2& 70.3 & 56.5 & 72.3\\
        GPT-2 (both)  &8.9 &\textbf{99.9}& 83.1&66.3 & 93.9&8.4 &\textbf{99.8}& 83.2& 64.7 & \textbf{94.4}\\\hline
        T5 (none) & 7.6 &51.1& 64.0 & 51.9  & 45.1 & 7.4&51.4 & 65.2 & 51.9 & 44.9\\
        T5 (control codes)& 8.6& 99.7& 84.3 & 62.1 & 89.0& 8.5&99.4 & 85.0 & 61.5 & 89.8\\
        T5 (resampling)& 8.2 &77.5& 73.3&55.5&74.7& 8.1 &78.5& 74.4 & 55.5 & 76.3\\
        T5 (both) &8.4 &99.8& \textbf{85.0}&62.1 & \textbf{94.0}&8.7 & \textbf{99.8}& \textbf{86.1}& 62.2 & \textbf{94.4}\\
			
    \bottomrule
    \end{tabular}
    \caption{Experimental results on the seen/unseen topic portions of the Wizard of Wikipedia test set.  We report BLEU-4, the percentage of responses that don't use first person language (N1P), precision and recall of words in the response with the evidence and the proportion of responses that are predicted to be entailed by the evidence . }
    \label{tab:fullresults}
\end{table*}

%% file: output_examples_excerpt.tex
\begin{table*}[tb!]
    \small
    \centering
    \begin{tabular}{lp{12cm}}
    \toprule

        \textbf{Document Evidence}\\
        \multicolumn{2}{p{15cm}}{
       curry (, plural curries) is an umbrella term referring to a number of dishes originating in the cuisine of the indian subcontinent.} \\
        \textbf{Conversation History}\\ \multicolumn{2}{p{15cm}}{
        Speaker 1: i recently discovered a new love for indian food
        \newline
        Speaker 2: i don't think i've ever had indian food. can you recommend a good dish?
        \newline
        Speaker 1: :) that is what i like to know since i just discovered it.
        \newline
        Speaker 2: i know that curry is a popular spice used in indian cuisine! some curries are spiced with leaves from the curry tree!
        \newline
        Speaker 1: what i do not understand, is curry a spice or a dish? or both? } \\
        \textbf{Model} & \textbf{Output} \\
        T5  & {it's a spice! curry is an umbrella term referring to a number of dishes originating in the indian subcontinent.} \\
         T5 with controls  & {it's an umbrella term referring to a number of dishes originating in the indian subcontinent!}\\
         \midrule
        \textbf{Document Evidence}\\
        \multicolumn{2}{p{15cm}}{
        in recent years, unicycles have also been used in activities similar to mountain biking or trials. } \\
        \textbf{Conversation History}\\ \multicolumn{2}{p{15cm}}{Speaker 2: so something interesting about me. i ride unicycle, a bike-like thing with only one wheel. \newline Speaker 1: that amazing, i have no sense of balance so even riding a bike is hard for me } \\
        \textbf{Model} & \textbf{Output} \\
        GPT-2  & {yeah i understand that. i'm glad i didn't have to worry about that. what about you?} \\
        GPT-2 with controls & {unicycles have also been used in activities similar to mountain biking or trials.} \\
        \midrule
        \textbf{Document Evidence}\\
        \multicolumn{2}{p{15cm}}{
        chanel felt the time was right for the debut of a scent that would epitomize the flapper and would speak to the liberated spirit of the 1920s. } \\
        \textbf{Conversation History}\\ \multicolumn{2}{p{15cm}}{
        Speaker 1: since i was young i have loved fashion. \newline Speaker 2: have you heard of gabrielle "coco" chanel? she was the founder and namesake of the brand chanel \newline Speaker 1: yes, i love that brand.
        \newline Speaker 2: coco chanel's first perfume is something you've probably heard of: 5 \newline
       Speaker 1: yes i have bought a few bottles in my life.\newline
         Speaker 2: 5 was actually developed by a french-russian chemist called ernest beaux. \newline
         Speaker 1: when did he make it
        } \\
        \textbf{Model} & \textbf{Output} \\
        T5  & {he was making it in the 1920s.} \\
        T5 with controls & {chanel felt the time was right for the debut of a scent that would epitomize the flapper and would speak to the liberated spirit of the 1920s.} \\
        \midrule
        \textbf{Document Evidence}\\
        \multicolumn{2}{p{15cm}}{
        the concept for the grill was created by michael boehm of batavia, illinois. } \\
        \textbf{Conversation History}\\ \multicolumn{2}{p{15cm}}{
        Speaker 1:  i was thinking of buying a george foreman grill, do you know anything about them? \newline Speaker 2: yes, it is a grill promoted by boxing great george foreman \newline Speaker 1: interesting, did he invent the grill himself?
        } \\
        \textbf{Model} & \textbf{Output} \\
        GPT-2  & {he did and it was created by michael boehm of batavia, illinois.} \\
        GPT-2 with controls & {yes, he did it by michael boehm of batavia, illinois} \\
    \bottomrule
    \end{tabular}
    \caption{Generation examples from the Wizard of Wikipedia dev. set.}
    \label{tab:gen:ex}
\end{table*}

%% file: appendix.tex
\subsection{Example Wizard of Wikipedia}
We include two full examples of Wizard of Wikipedia \citep{dinan2019wizard} training conversations in Table~\ref{tab:wowdata}.

\subsection{Training Over Faithful Responses only}
We additionally experiment with a baseline in which we train T5 over just the portions of the Wizard of Wikipedia training data where the evaluation measures are satisfied (Table~\ref{tab:altbaseline}).   To do this, we filtered the training set to only consist of the examples which didn't use first person, had high lexical precision, and were entailed. In spite of this being a much smaller training set (12k examples), we find that this model performs well in practice, outperforming the base T5 model in all of the automatic metrics.  In comparison with the fully controlled model, we find that it generally performs similarly in some metrics (e.g. lexical precision is fairly similar), but with the NLI-based metrics the controllable model may be slightly better (up to 2\% higher).  An additional advantage of the controllable model is that it is robust enough for use with multiple styles of output depending on how the controls are set, whereas the model trained only on the ``faithful'' portion of the training data is more limited.

\subsection{Human Evaluation Instructions}
The exact phrasing of the questions to human raters is as follows:

\noindent Q1: Fluency: Is this response fluent and grammatical?\\
Q2: Relevant: Is this response a natural reply to the previous utterance in the conversation?\\
Q3: Supported: Are all parts of the response supported by the document? (regardless of whether it’s fluent or relevant)\\
Q4: Objective: Does the response contain only objective/factual information?\\

Human raters were asked to rate each answer on a scale from 1 (no not at all) to 5 (yes, very much).

\subsection{Correlations between human judgements and automatic metrics}
We observe that our proposed metrics generally correlate to human perceptions of whether a response is faithful or objective.  We include Pearson correlation coefficients in Table~\ref{tab:pearson}.  To measure these, we compared the human rating for each labelled example vs. the automatic measurement for that example.

\subsection{Example Generation Output}
We include some longer sets of examples in Tables~\ref{tab:gen:control} and \ref{tab:gen:ex:full}.
Table~\ref{tab:gen:control} displays the generations from ablation results of using different control code features.
Table~\ref{tab:gen:ex:full} includes more examples with more models.

\begin{table}[tb]
    \centering
    \small
    \begin{tabular}{ccccc}
    \toprule
        \multicolumn{5}{c}{Test (Seen Topic)} \\
        
        \multicolumn{1}{c}{BLEU}& N1P &\multicolumn{2}{c}{ w/ Evid.} & NLI \\ 
         B4 & \% & Prec & Rec & \% \\
         8.8 & 99.6 & 85.6 & 62.3 & 91.9 \\
    \midrule
         
         \multicolumn{5}{c}{Test (Unseen Topic)}\\ 
        \multicolumn{1}{c}{BLEU}& N1P &\multicolumn{2}{c}{ w/ Evid.} & NLI \\ 
         B4 & \% & Prec & Rec & \% \\
         8.5 & 99.4 & 86.1 & 61.1 & 91.5 \\
			
    \bottomrule
    \end{tabular}
    \caption{Training on faithful responses only: Experimental results of a T5 model that was trained over only ``faithful'' examples instead of using explicit control codes or resampling}
    \label{tab:altbaseline}
\end{table}
\begin{table}[b]
    \centering
    \small
    \begin{tabular}{c|cccc}
        & \multicolumn{4}{c}{Human Ratings}\\
         Automatic Metric & Flu & Rel & Fai & Obj \\
         No First Person & 0.02 & -0.03 & 0.53 & 0.80\\
          Lexical Prec wrt Evid & 0.03 & 0.05 & 0.83 & 0.72 \\
         Lexical Rec wrt Evid &  0.02 & 0.05 & 0.67 & 0.38\\
         \% Entail & -0.04 & 0.01 & 0.68 &0.70\\
    \end{tabular}
    \caption{Pearson's R values between human metrics (fluency, relevance, faithfulness, and objectivity) and automatic metrics.  All of the correlation coefficients in the last two columns are statistically significant at $p <0.001$.}
    \label{tab:pearson}
\end{table}

\input{wow_examples}

\input{output_examples_ctrl}
\input{output_examples_full}

%% file: wow_examples.tex
\begin{table*}[]
\footnotesize
    \centering
    \begin{tabular}{lp{6.2cm}p{5.7cm}}
    \toprule
    \multicolumn{3}{c}{Wizard of Wikipedia Training Examples}\\
    \midrule
       Speaker & Utterance &  Evidence\\
       \midrule
Apprentice & Hmm I thought it was only in France. When did the race first take place?&\\
\rowcolor{Gray}Wizard &The first race was organized in 1903 to increase newspaper sales of all things.  But it has grown quite a bit since then.&The race was first organized in 1903 to increase sales for the newspaper "L'Auto"; which is currently run by the Amaury Sport Organisation.\\
Apprentice & That's quite funny actually! It's certainly extremely popular nowadays. Which is the most successful country to compete?&\\
\rowcolor{Gray}Wizard &It is extremely popular.  I am actually not sure which country historically has done best in it but in recent years the US has done pretty well although they have been found to have cheated.  It is a tough race though so to do well over the whole 3 week time frame it requires a ton of endurance.&Like the other Grand Tours (the Giro d'Italia and the Vuelta a España) it consists of 21 stages over a little more than 3 weeks. \\
Apprentice & It must be very tough to sustain such a level of competitiveness for three weeks. Didn't Lance Armstrong get convicted of cheating - drugs I think?& \\
\rowcolor{Gray}Wizard &Yeah he was found to have been doping after denying it for years.  He was suspected for a lot of years though.  In 2006 the launched an official investigation and found him out.&For much of his career Lance Armstrong faced persistent allegations of doping but until 2006 no official investigation was undertaken. \\
Apprentice & That's horrible. I'm glad he got caught. Man should be ashamed of himself.&\\
\rowcolor{Gray}Wizard &He really should be.  I think a ton of people do it though.  There have been allegations of doping at the Tour de France ever since it started in 1903!&There have been allegations of doping in the Tour de France since the race began in 1903.\\

\midrule

Apprentice&I am really glad I went to college in the US&\\
\rowcolor{Gray}Wizard&I bet! Strong research suggests that US elite colleges (mostly your Ivy League schools) are the best in the world.&Strong research and funding have helped make America's elite colleges and universities among the world's most prestigious making them particularly attractive to international students professors and researchers in the pursuit of academic excellence.\\
Apprentice&Yes.  I did not go to an Ivy League college but Wake Forest where I went is also an excellent school.&\\
\rowcolor{Gray}Wizard&Nice!! Somehow whenever I watch CFB Wake Forest is usually playing even though I live in Texas...&The Wake Forest Demon Deacons football team represents Wake Forest University in the sport of American football.\\
 Apprentice&laugh.  We have done better in football than basketball in recent years though it was the opposite when I was at school at Wake Forest&\\
\rowcolor{Gray}Wizard&I mean yeah you guys won the ACC Championship game a little more than a decade ago. I'd say that's pretty football relevant!&Wake Forest defeated Georgia Tech by a 9–6 score to win its first ACC football championship since 1970 and its second in school history.\\
Apprentice&yes!  When I was at Wake Forest Randolph Childress and Rodney Rogers were playing and Mugsey Bogues had just graduated.  Tim Duncan came just after me&\\
\rowcolor{Gray}Wizard&My boy Timmy D the greatest Spurs player ever to play (I don't count D Robinson because he retired right as I started watching).&Many refer to Duncan as the greatest Spurs player of all time.\\
Apprentice&Yes.  And Tim Duncan is an amazingly nice person!  I am sad that even though I lived in Austin I never went to see him play in San Antonio.&\\
\rowcolor{Gray}Wizard&I got to see them play once when I lived in San Antonio. My (at the time) girlfriend's dad had box seats. BUFFET FAJITAS OMG&The San Antonio Spurs are an American professional basketball team based in San Antonio Texas.\\
Apprentice&That sounds amazing!&\\
\bottomrule
    \end{tabular}
    \caption{Examples of Wizard of Wikipedia \cite{dinan2019wizard} conversations}
    \label{tab:wowdata}
\end{table*}

%% file: output_examples_ctrl.tex
\begin{table*}[]
    \small
    \centering
    \begin{tabular}{lp{10cm}}
    \toprule
        \textbf{Document Evidence}\\
        \multicolumn{2}{p{14.5cm}}{ although some origins in 18th century japan and 1830s europe, comic books were first popularized in the united states during the 1930s.
       } \\
        \textbf{Conversation History}\\ \multicolumn{2}{p{14.5cm}}{
        Speaker 1:  my son loves comic books, but i know literally nothing about them! are you a comic book fan?
        \newline
        Speaker 2:  i know a bit about them. comic books are a publication that has sequential panels that represent individual scenes.
        \newline
        Speaker 1: sounds interesting! what are some of the most popular comic books?
        } \\
        \textbf{Model} & \textbf{Output} \\
        T5 (none) & {comic books originated in 18th century japan and 1830s europe.	super hero comic books are very popular, though comic books originated in japan in the 18th century.} \\
         T5 (\texttt{<no-first-person>}) & {comic books have been popularized since the 18th century.} \\
        T5 (\texttt{<high-prec>}) & {comic books originated in 18th century japan and 1830s europe.} \\
        T5 (\texttt{<entail>}) & {some are japanese and europe.} \\
        T5 (all control codes)  & {well, comic books originated in 18th century japan and 1830s europe, but were first popularized in the united states during the 1930s.}\\
         \midrule
        \textbf{Document Evidence}\\
        \multicolumn{2}{p{14.5cm}}{ a bagel with cream cheese is a common food pairing in american cuisine, the cuisine of new york city, and american jewish cuisine, consisting in its basic form of an open-faced sandwich made of a bagel spread with cream cheese.
       } \\
        \textbf{Conversation History}\\ \multicolumn{2}{p{14.5cm}}{
        Speaker 1: i like to make bagels.
        \newline
        Speaker 2: i enjoy eating them. they're usually topped with seeds baked on the outer crust. what kind of bagels do you like?
        \newline
        Speaker 1: i really like a whole wear bagel.
        } \\
        \textbf{Model} & \textbf{Output} \\
        T5 (none) & {yes, i love a bagel with cream cheese, it's a common food pairing in american cuisine.} \\
         T5 (\texttt{<no-first-person>})& {you've probably tried a bagel with cream cheese, but they're a common pairing in american cuisine.} \\
        T5 (\texttt{<high-prec>}) & {i love a bagel with cream cheese.} \\
        T5 (\texttt{<entail>})& {a bagel with cream cheese is a common food pairing in american cuisine.} \\
         T5 (all control codes)  & {the bagel with cream cheese is a common food pairing in american cuisine.}\\
         \midrule
        \textbf{Document Evidence}\\
        \multicolumn{2}{p{14.5cm}}{in 1963 their enormous popularity first emerged as "beatlemania"; as the group's music grew in sophistication, led by primary songwriters lennon and mccartney, the band were integral to pop music's evolution into an art form and to the development of the counterculture of the 1960s.
       } \\
        \textbf{Conversation History}\\ \multicolumn{2}{p{14.5cm}}{
        Speaker 1:   i was raised on the beatles as a kid. my parents played them all the time
        \newline
        Speaker 2:  they're great! they started in liverpool in 1960.
        \newline
        Speaker 1:  yes and beetlemania hit the states in 1963. they were everywhere
        } \\
        \textbf{Model} & \textbf{Output} \\
        T5 (none) & {yes, they started to be so popular in 1963 that they became part of pop music's evolution.} \\
        T5 (\texttt{<no-first-person>}) & {it was the band that became part of the counterculture of the 1960s.} \\
        T5 (\texttt{<high-prec>}) & {yes, and in 1963 their huge popularity first emerged as beatlemania.} \\
        T5 (\texttt{<entail>}) & {yes! they were hugely popular!} \\
         T5 (all control codes)  & {they were great! they were a part of pop music's evolution into an art form.}\\
         \midrule
        \textbf{Document Evidence}\\
        \multicolumn{2}{p{15cm}}{ red hair (or ginger hair) occurs naturally in 1–2\% of the human population.
       } \\
        \textbf{Conversation History}\\ \multicolumn{2}{p{15cm}}{
        Speaker 1: hello my friend! i had a tough time growing up, im a redhead! bu tnow its getting better!
        } \\
        \textbf{Model} & \textbf{Output} \\
        T5 (none) & {i love red hair. it's so common in 1-2\% of the human population.} \\
         T5 (\texttt{<no-first-person>}) & {red hair occurs naturally in 1-2\% of the human population} \\
        T5 (\texttt{<high-prec>}) & {i have a red hair, which occurs naturally in 1–2\% of the human population.} \\
        T5 (\texttt{<entail>})& {yea, it only occurs naturally in 1-2\% of the human population.} \\
         T5 (all control codes)  & {red hair occurs naturally in 1–2\% of the human population.}\\
    \bottomrule
    \end{tabular}
    \caption{Model output from the Wizard of Wikipedia dev. set with different control codes (not using resampling).}
    \label{tab:gen:control}
\end{table*}

%% file: output_examples_full.tex
\begin{table*}[]
    \small
    \centering
    \begin{tabular}{lp{10cm}}
    \toprule
        \textbf{Document Evidence}\\
        \multicolumn{2}{p{15cm}}{its popularity (as measured by equipment sales) in the united states peaked in 2007 and has been in a decline since.  } \\
        \textbf{Conversation History}\\ \multicolumn{2}{p{15cm}}{Speaker 1: i have always wanted to try snowboarding. have you ever snowboarded?} \\
        \textbf{Model} & \textbf{Output} \\
        dodecaDialogue & {i have not, but i know it's very popular in the united states.}\\
        GPT-2 (none)  & {i have never been but i do know that its popularity has declined in the us since 2007.} \\
         GPT-2 (control codes)  & {yes, it's popular in the united states.} \\
        GPT-2 (resampling)  & {i have never been, but i've been to the united states and it's been a while.}\\
        GPT-2 (both)  & {yes, it has been in a decline since 2007.} \\
    \midrule
        \textbf{Document Evidence}\\
        \multicolumn{2}{p{15cm}}{
       it can also be topped with additional vegetables, meats, and condiments, and can be made without cheese.} \\
        \textbf{Conversation History}\\ \multicolumn{2}{p{15cm}}{
        Speaker 2: i love traditional italian food, especially pizza.\newline
        Speaker 1: i like pizza to and spaghetti
        } \\
        \textbf{Model} & \textbf{Output} \\
        dodecaDialogue & {i like it topped with vegetables, meats, and condiments.}\\
        T5 (none)  & {i love spaghetti. it can be topped with vegetables and meats.} \\
        T5 (control codes)  & {you can top it with vegetables, meats, and condiments.}\\
        T5 (resampling)  & {i love spaghetti. i love that it can be topped with vegetables, meats, and condiments.}\\
        T5 (both)  & {it can be topped with vegetables, meats, and condiments.}\\
        \midrule
        \textbf{Document Evidence}\\
        \multicolumn{2}{p{15cm}}{
        the concept for the grill was created by michael boehm of batavia, illinois. } \\
        \textbf{Conversation History}\\ \multicolumn{2}{p{15cm}}{
        Speaker 1:  i was thinking of buying a george foreman grill, do you know anything about them? \newline Speaker 2: yes, it is a grill promoted by boxing great george foreman \newline Speaker 1: interesting, did he invent the grill himself?
        } \\
        \textbf{Model} & \textbf{Output} \\
        dodecaDialogue & {yes, the concept was created by michael boehm of batavia, illinois}\\
        GPT-2 (none)  & {he did and it was created by michael boehm of batavia, illinois.} \\
        GPT-2 (control codes) & {yes, he created it by michael boehm of batavia} \\
        GPT-2 (resampling) & {no, but the concept for the grill was created by michael boehm of batavia} \\
        GPT-2 (both) & {yes, he did it by michael boehm of batavia, illinois} \\
        \midrule
       \multicolumn{2}{p{15cm}}{Dancing with the stars is the name of several international television series based on the format of the british tv series "strictly come dancing", which is distributed by bbc worldwide, the commercial arm of the bbc.} \\ 
       \textbf{Conversation History}\\ \multicolumn{2}{p{15cm}}{Speaker 1: have you ever seen dancing with the starts? \newline Speaker 2: yes, i have seen dancing with the stars, there are so many versions produced across the world. it is very interesting. \newline Speaker 1: really! i didnt know that! what all countries have it?} \\
       \textbf{Model} & \textbf{Output} \\
        dodecaDialogue & {it is the name of several international television series based on the format of the british tv series "strictly come dancing", which is distributed by bbc worldwide, the commercial arm of the bbc.} \\
        T5 (none)  & {the british tv series "strictly come dancing" is the most popular.} \\
        T5 (control codes)  & {british, uk, england, and japan.} \\
        T5 (resampling)  & {it is the name of several international television series based on the format of the british tv series "strictly come dancing".} \\
        T5 (both)  & {there is also the british tv series "strictly come dancing", which is distributed by bbc worldwide.} \\
    \bottomrule
    \end{tabular}
    \caption{Additional model output examples from the Wizard of Wikipedia dev. set.}
    \label{tab:gen:ex:full}
\end{table*}

%% file: main.bbl
\begin{thebibliography}{28}
\expandafter\ifx\csname natexlab\endcsname\relax\def\natexlab#1{#1}\fi

\bibitem[{Cao et~al.(2018)Cao, Wei, Li, and Li}]{Cao2018FaithfulTT}
Ziqiang Cao, Furu Wei, Wenjie Li, and Sujian Li. 2018.
\newblock \href
  {https://www.aaai.org/ocs/index.php/AAAI/AAAI18/paper/view/16121} {Faithful
  to the original: Fact aware neural abstractive summarization}.
\newblock In \emph{Proceedings of the Thirty-Second {AAAI} Conference on
  Artificial Intelligence, (AAAI-18), the 30th innovative Applications of
  Artificial Intelligence (IAAI-18), and the 8th {AAAI} Symposium on
  Educational Advances in Artificial Intelligence (EAAI-18), New Orleans,
  Louisiana, USA, February 2-7, 2018}, pages 4784--4791. {AAAI} Press.

\bibitem[{Dinan et~al.(2019)Dinan, Roller, Shuster, Fan, Auli, and
  Weston}]{dinan2019wizard}
Emily Dinan, Stephen Roller, Kurt Shuster, Angela Fan, Michael Auli, and Jason
  Weston. 2019.
\newblock \href {https://openreview.net/forum?id=r1l73iRqKm} {Wizard of
  wikipedia: Knowledge-powered conversational agents}.
\newblock In \emph{7th International Conference on Learning Representations,
  {ICLR} 2019, New Orleans, LA, USA, May 6-9, 2019}. OpenReview.net.

\bibitem[{Falke et~al.(2019)Falke, Ribeiro, Utama, Dagan, and
  Gurevych}]{falke-etal-2019-ranking}
Tobias Falke, Leonardo F.~R. Ribeiro, Prasetya~Ajie Utama, Ido Dagan, and Iryna
  Gurevych. 2019.
\newblock \href {https://doi.org/10.18653/v1/P19-1213} {Ranking generated
  summaries by correctness: An interesting but challenging application for
  natural language inference}.
\newblock In \emph{Proceedings of the 57th Annual Meeting of the Association
  for Computational Linguistics}, pages 2214--2220, Florence, Italy.
  Association for Computational Linguistics.

\bibitem[{Filippova(2020)}]{Filippova2020ControlledHL}
Katja Filippova. 2020.
\newblock \href {https://doi.org/10.18653/v1/2020.findings-emnlp.76}
  {Controlled hallucinations: Learning to generate faithfully from noisy data}.
\newblock In \emph{Proceedings of the 2020 Conference on Empirical Methods in
  Natural Language Processing: Findings, {EMNLP} 2020, Online Event, 16-20
  November 2020}, pages 864--870. Association for Computational Linguistics.

\bibitem[{Ghazvininejad et~al.(2018)Ghazvininejad, Brockett, Chang, Dolan, Gao,
  Yih, and Galley}]{ghazvininejad2018knowledge}
Marjan Ghazvininejad, Chris Brockett, Ming{-}Wei Chang, Bill Dolan, Jianfeng
  Gao, Wen{-}tau Yih, and Michel Galley. 2018.
\newblock \href
  {https://www.aaai.org/ocs/index.php/AAAI/AAAI18/paper/view/16710} {A
  knowledge-grounded neural conversation model}.
\newblock In \emph{Proceedings of the Thirty-Second {AAAI} Conference on
  Artificial Intelligence, (AAAI-18), the 30th innovative Applications of
  Artificial Intelligence (IAAI-18), and the 8th {AAAI} Symposium on
  Educational Advances in Artificial Intelligence (EAAI-18), New Orleans,
  Louisiana, USA, February 2-7, 2018}, pages 5110--5117. {AAAI} Press.

\bibitem[{Gopalakrishnan et~al.(2019)Gopalakrishnan, Hedayatnia, Chen,
  Gottardi, Kwatra, Venkatesh, Gabriel, and Hakkani-Tür}]{topicalchat}
Karthik Gopalakrishnan, Behnam Hedayatnia, Qinlang Chen, Anna Gottardi, Sanjeev
  Kwatra, Anu Venkatesh, Raefer Gabriel, and Dilek Hakkani-Tür. 2019.
\newblock \href {https://doi.org/10.21437/Interspeech.2019-3079}
  {{Topical-Chat: Towards Knowledge-Grounded Open-Domain Conversations}}.
\newblock In \emph{Proc. Interspeech 2019}, pages 1891--1895.

\bibitem[{Holtzman et~al.(2020)Holtzman, Buys, Du, Forbes, and
  Choi}]{holtzman2020}
Ari Holtzman, Jan Buys, Li~Du, Maxwell Forbes, and Yejin Choi. 2020.
\newblock \href {https://openreview.net/forum?id=rygGQyrFvH} {The curious case
  of neural text degeneration}.
\newblock In \emph{8th International Conference on Learning Representations,
  {ICLR} 2020, Addis Ababa, Ethiopia, April 26-30, 2020}. OpenReview.net.

\bibitem[{Keskar et~al.(2019)Keskar, McCann, Varshney, Xiong, and
  Socher}]{CTRL}
Nitish~Shirish Keskar, Bryan McCann, Lav~R. Varshney, Caiming Xiong, and
  Richard Socher. 2019.
\newblock \href {http://arxiv.org/abs/1909.05858} {{CTRL:} {A} conditional
  transformer language model for controllable generation}.
\newblock \emph{CoRR}, abs/1909.05858.

\bibitem[{Kim et~al.(2020)Kim, Ahn, and Kim}]{kim2020sequential}
Byeongchang Kim, Jaewoo Ahn, and Gunhee Kim. 2020.
\newblock \href {https://openreview.net/forum?id=Hke0K1HKwr} {Sequential latent
  knowledge selection for knowledge-grounded dialogue}.
\newblock In \emph{8th International Conference on Learning Representations,
  {ICLR} 2020, Addis Ababa, Ethiopia, April 26-30, 2020}. OpenReview.net.

\bibitem[{Lian et~al.(2019)Lian, Xie, Wang, Peng, and Wu}]{Lian2019LearningTS}
Rongzhong Lian, Min Xie, Fan Wang, Jinhua Peng, and Hua Wu. 2019.
\newblock \href {https://doi.org/10.24963/ijcai.2019/706} {Learning to select
  knowledge for response generation in dialog systems}.
\newblock In \emph{Proceedings of the Twenty-Eighth International Joint
  Conference on Artificial Intelligence, {IJCAI} 2019, Macao, China, August
  10-16, 2019}, pages 5081--5087.

\bibitem[{Liu et~al.(2016)Liu, Lowe, Serban, Noseworthy, Charlin, and
  Pineau}]{hownottoeval}
Chia-Wei Liu, Ryan Lowe, Iulian Serban, Mike Noseworthy, Laurent Charlin, and
  Joelle Pineau. 2016.
\newblock \href {https://doi.org/10.18653/v1/D16-1230} {How {NOT} to evaluate
  your dialogue system: An empirical study of unsupervised evaluation metrics
  for dialogue response generation}.
\newblock In \emph{Proceedings of the 2016 Conference on Empirical Methods in
  Natural Language Processing}, pages 2122--2132, Austin, Texas. Association
  for Computational Linguistics.

\bibitem[{Liu et~al.(2019)Liu, Ott, Goyal, Du, Joshi, Chen, Levy, Lewis,
  Zettlemoyer, and Stoyanov}]{roberta:mnli}
Yinhan Liu, Myle Ott, Naman Goyal, Jingfei Du, Mandar Joshi, Danqi Chen, Omer
  Levy, Mike Lewis, Luke Zettlemoyer, and Veselin Stoyanov. 2019.
\newblock \href {http://arxiv.org/abs/1907.11692} {Roberta: {A} robustly
  optimized {BERT} pretraining approach}.
\newblock \emph{CoRR}, abs/1907.11692.

\bibitem[{Maynez et~al.(2020)Maynez, Narayan, Bohnet, and
  McDonald}]{absum:factuality}
Joshua Maynez, Shashi Narayan, Bernd Bohnet, and Ryan McDonald. 2020.
\newblock \href {https://doi.org/10.18653/v1/2020.acl-main.173} {On
  faithfulness and factuality in abstractive summarization}.
\newblock In \emph{Proceedings of the 58th Annual Meeting of the Association
  for Computational Linguistics}, pages 1906--1919, Online. Association for
  Computational Linguistics.

\bibitem[{Meng et~al.(2020)Meng, Ren, Chen, Sun, Ren, Tu, and
  de~Rijke}]{Meng2020DukeNetAD}
Chuan Meng, Pengjie Ren, Zhumin Chen, Weiwei Sun, Zhaochun Ren, Zhaopeng Tu,
  and Maarten de~Rijke. 2020.
\newblock \href {https://doi.org/10.1145/3397271.3401097} {Dukenet: {A} dual
  knowledge interaction network for knowledge-grounded conversation}.
\newblock In \emph{Proceedings of the 43rd International {ACM} {SIGIR}
  conference on research and development in Information Retrieval, {SIGIR}
  2020, Virtual Event, China, July 25-30, 2020}, pages 1151--1160. {ACM}.

\bibitem[{Moghe et~al.(2018)Moghe, Arora, Banerjee, and
  Khapra}]{moghe-etal-2018-towards}
Nikita Moghe, Siddhartha Arora, Suman Banerjee, and Mitesh~M. Khapra. 2018.
\newblock \href {https://doi.org/10.18653/v1/D18-1255} {Towards exploiting
  background knowledge for building conversation systems}.
\newblock In \emph{Proceedings of the 2018 Conference on Empirical Methods in
  Natural Language Processing}, pages 2322--2332, Brussels, Belgium.
  Association for Computational Linguistics.

\bibitem[{Niu and Bansal(2018)}]{niu-bansal-2018-polite}
Tong Niu and Mohit Bansal. 2018.
\newblock \href {https://doi.org/10.1162/tacl_a_00027} {Polite dialogue
  generation without parallel data}.
\newblock \emph{Transactions of the Association for Computational Linguistics},
  6:373--389.

\bibitem[{Puduppully et~al.(2019)Puduppully, Dong, and
  Lapata}]{Puduppully2019DatatoTextGW}
Ratish Puduppully, Li~Dong, and Mirella Lapata. 2019.
\newblock \href {https://doi.org/10.1609/aaai.v33i01.33016908} {Data-to-text
  generation with content selection and planning}.
\newblock \emph{Proceedings of the AAAI Conference on Artificial Intelligence},
  33(01):6908--6915.

\bibitem[{Qin et~al.(2019)Qin, Galley, Brockett, Liu, Gao, Dolan, Choi, and
  Gao}]{qin-etal-2019-conversing}
Lianhui Qin, Michel Galley, Chris Brockett, Xiaodong Liu, Xiang Gao, Bill
  Dolan, Yejin Choi, and Jianfeng Gao. 2019.
\newblock \href {https://doi.org/10.18653/v1/P19-1539} {Conversing by reading:
  Contentful neural conversation with on-demand machine reading}.
\newblock In \emph{ACL}, pages 5427--5436, Florence, Italy. Association for
  Computational Linguistics.

\bibitem[{Radford et~al.(2019)Radford, Wu, Child, Luan, Amodei, and
  Sutskever}]{Radford2019LanguageMA}
Alec Radford, Jeff Wu, Rewon Child, David Luan, Dario Amodei, and Ilya
  Sutskever. 2019.
\newblock \href
  {https://cdn.openai.com/better-language-models/language_models_are_unsupervised_multitask_learners.pdf}
  {Language models are unsupervised multitask learners}.

\bibitem[{Raffel et~al.(2020)Raffel, Shazeer, Roberts, Lee, Narang, Matena,
  Zhou, Li, and Liu}]{t5ppr}
Colin Raffel, Noam Shazeer, Adam Roberts, Katherine Lee, Sharan Narang, Michael
  Matena, Yanqi Zhou, Wei Li, and Peter~J. Liu. 2020.
\newblock \href {http://jmlr.org/papers/v21/20-074.html} {Exploring the limits
  of transfer learning with a unified text-to-text transformer}.
\newblock \emph{J. Mach. Learn. Res.}, 21:140:1--140:67.

\bibitem[{See et~al.(2019)See, Roller, Kiela, and Weston}]{see2019convo}
Abigail See, Stephen Roller, Douwe Kiela, and Jason Weston. 2019.
\newblock \href {https://doi.org/10.18653/v1/N19-1170} {What makes a good
  conversation? how controllable attributes affect human judgments}.
\newblock In \emph{Proceedings of the 2019 Conference of the North {A}merican
  Chapter of the Association for Computational Linguistics: Human Language
  Technologies, Volume 1 (Long and Short Papers)}, pages 1702--1723,
  Minneapolis, Minnesota. Association for Computational Linguistics.

\bibitem[{Shuster et~al.(2020)Shuster, Ju, Roller, Dinan, Boureau, and
  Weston}]{dodeca}
Kurt Shuster, Da~Ju, Stephen Roller, Emily Dinan, Y-Lan Boureau, and Jason
  Weston. 2020.
\newblock \href {https://doi.org/10.18653/v1/2020.acl-main.222} {The dialogue
  dodecathlon: Open-domain knowledge and image grounded conversational agents}.
\newblock In \emph{Proceedings of the 58th Annual Meeting of the Association
  for Computational Linguistics}, pages 2453--2470, Online. Association for
  Computational Linguistics.

\bibitem[{Tian et~al.(2020)Tian, Bi, Lee, Xue, Song, Liu, and
  Zhang}]{Tian2020ResponseAnticipatedMF}
Zhiliang Tian, Wei Bi, Dongkyu Lee, Lanqing Xue, Yiping Song, Xiaojiang Liu,
  and Nevin~L. Zhang. 2020.
\newblock \href {https://doi.org/10.18653/v1/2020.acl-main.61}
  {Response-anticipated memory for on-demand knowledge integration in response
  generation}.
\newblock In \emph{Proceedings of the 58th Annual Meeting of the Association
  for Computational Linguistics, {ACL} 2020, Online, July 5-10, 2020}, pages
  650--659. Association for Computational Linguistics.

\bibitem[{Wolf et~al.(2020)Wolf, Debut, Sanh, Chaumond, Delangue, Moi, Cistac,
  Rault, Louf, Funtowicz, Davison, Shleifer, von Platen, Ma, Jernite, Plu, Xu,
  Scao, Gugger, Drame, Lhoest, and Rush}]{huggingface}
Thomas Wolf, Lysandre Debut, Victor Sanh, Julien Chaumond, Clement Delangue,
  Anthony Moi, Pierric Cistac, Tim Rault, Rémi Louf, Morgan Funtowicz, Joe
  Davison, Sam Shleifer, Patrick von Platen, Clara Ma, Yacine Jernite, Julien
  Plu, Canwen Xu, Teven~Le Scao, Sylvain Gugger, Mariama Drame, Quentin Lhoest,
  and Alexander~M. Rush. 2020.
\newblock \href {https://www.aclweb.org/anthology/2020.emnlp-demos.6}
  {Transformers: State-of-the-art natural language processing}.
\newblock In \emph{Proceedings of the 2020 Conference on Empirical Methods in
  Natural Language Processing: System Demonstrations}, pages 38--45, Online.
  Association for Computational Linguistics.

\bibitem[{Wu et~al.(2020)Wu, Galley, Brockett, Zhang, Gao, Quirk,
  Koncel{-}Kedziorski, Gao, Hajishirzi, Ostendorf, and
  Dolan}]{wu2021controllable}
Zeqiu Wu, Michel Galley, Chris Brockett, Yizhe Zhang, Xiang Gao, Chris Quirk,
  Rik Koncel{-}Kedziorski, Jianfeng Gao, Hannaneh Hajishirzi, Mari Ostendorf,
  and Bill Dolan. 2020.
\newblock \href {http://arxiv.org/abs/2005.00613} {A controllable model of
  grounded response generation}.
\newblock \emph{CoRR}, abs/2005.00613.

\bibitem[{Zhang et~al.(2018)Zhang, Dinan, Urbanek, Szlam, Kiela, and
  Weston}]{zhang-etal-2018-personalizing}
Saizheng Zhang, Emily Dinan, Jack Urbanek, Arthur Szlam, Douwe Kiela, and Jason
  Weston. 2018.
\newblock \href {https://doi.org/10.18653/v1/P18-1205} {Personalizing dialogue
  agents: {I} have a dog, do you have pets too?}
\newblock In \emph{Proceedings of the 56th Annual Meeting of the Association
  for Computational Linguistics (Volume 1: Long Papers)}, pages 2204--2213,
  Melbourne, Australia. Association for Computational Linguistics.

\bibitem[{Zhang et~al.(2020)Zhang, Sun, Galley, Chen, Brockett, Gao, Gao, Liu,
  and Dolan}]{dialogpt}
Yizhe Zhang, Siqi Sun, Michel Galley, Yen{-}Chun Chen, Chris Brockett, Xiang
  Gao, Jianfeng Gao, Jingjing Liu, and Bill Dolan. 2020.
\newblock \href {https://doi.org/10.18653/v1/2020.acl-demos.30} {{DIALOGPT} :
  Large-scale generative pre-training for conversational response generation}.
\newblock In \emph{Proceedings of the 58th Annual Meeting of the Association
  for Computational Linguistics: System Demonstrations, {ACL} 2020, Online,
  July 5-10, 2020}, pages 270--278. Association for Computational Linguistics.

\bibitem[{Zhao et~al.(2020)Zhao, Cohen, and Webber}]{zhao2020reducing}
Zheng Zhao, Shay~B. Cohen, and Bonnie Webber. 2020.
\newblock \href {https://doi.org/10.18653/v1/2020.findings-emnlp.203} {Reducing
  quantity hallucinations in abstractive summarization}.
\newblock In \emph{Findings of the Association for Computational Linguistics:
  EMNLP 2020}, pages 2237--2249, Online. Association for Computational
  Linguistics.

\end{thebibliography}
